\documentclass{article}
\usepackage{PRIMEarxiv}

\usepackage[utf8]{inputenc}
\usepackage[T1]{fontenc}
\usepackage{hyperref}
\usepackage{url}
\usepackage{booktabs}
\usepackage{amsfonts}
\usepackage{amssymb}
\usepackage{nicefrac}
\usepackage{microtype}
\usepackage{graphicx}
\usepackage{multirow}
\usepackage{array}
\usepackage{siunitx}
\usepackage{xcolor}
\usepackage{listings}
\usepackage{natbib}

\definecolor{codegreen}{rgb}{0,0.6,0}
\definecolor{codegray}{rgb}{0.5,0.5,0.5}
\definecolor{codepurple}{rgb}{0.58,0,0.82}
\definecolor{backcolour}{rgb}{0.97,0.97,0.97}

\usepackage[table]{xcolor}

\usepackage{microtype}
\microtypesetup{stretch=15, shrink=15, protrusion=true, expansion=true}
\usepackage{graphicx}
\usepackage{booktabs}
\usepackage{enumitem}
\usepackage{amsmath}
\usepackage{amsthm}         
\theoremstyle{definition}
\newtheorem{definition}{Definition}
\usepackage{hyperref}       
\usepackage{cleveref}
\usepackage{tikz}
\usetikzlibrary{arrows.meta,positioning,calc,patterns}
\usepackage{pgffor}     
\usepackage{pgfplots}
\pgfplotsset{compat=1.18}

\definecolor{cellred}{RGB}{240,150,150}
\definecolor{cellamber}{RGB}{255,210,100}
\definecolor{cellgreen}{RGB}{180,230,180}
\newcommand{\rcell}{\cellcolor{cellred}}
\newcommand{\acell}{\cellcolor{cellamber}}
\newcommand{\gcell}{\cellcolor{cellgreen}}

\definecolor{covweak}{RGB}{240,240,240}
\definecolor{covmed}{RGB}{180,220,180}
\definecolor{covstrong}{RGB}{80,180,80}
\definecolor{covvstrong}{RGB}{20,120,20}
\DeclareRobustCommand{\dotscale}[1]{%
  \foreach \i in {1,...,4}{%
    \ifnum\i>#1 $\circ$\else$\bullet$\fi\,%
  }%
}
\title{\includegraphics[width=0.99\textwidth]{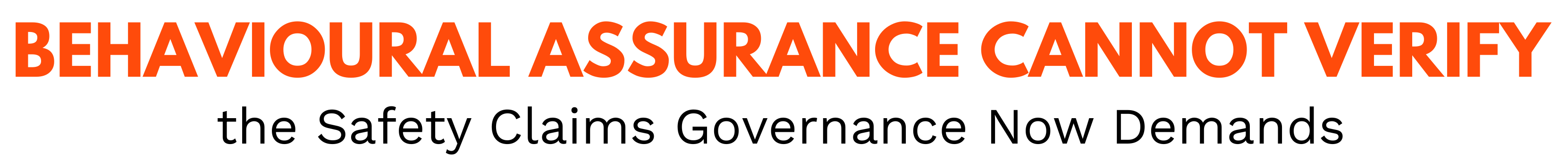}}

\author{
  Pratinav Seth, Vinay Kumar Sankarapu \\
  \affiliation{Lexsi Labs}\\
  pratinav.seth@lexsi.ai \\
}

\setabstract{%
This position paper argues that behavioural assurance, even when carefully designed, is being asked to carry safety claims it cannot verify. AI governance frameworks enacted between 2019 and early 2026 require reviewable evidence of properties such as the absence of hidden objectives, resistance to loss-of-control precursors, and bounded catastrophic capability; current assurance methodologies (primarily behavioural evaluations and red-teaming) are epistemically limited to observable model outputs and cannot verify the latent representations or long-horizon agentic behaviours these frameworks presume to regulate. 
We formalize this structural mismatch as the \emph{audit gap}, the divergence between required and achievable verification access, and introduce the concept of \emph{fragile assurance} to describe cases where the evidential structure does not support the asserted safety claim. Through an analysis of a 21-instrument inventory, we identify an \emph{incentive gradient} where geopolitical and industrial pressures systematically reward surface-level behavioral proxies over deep structural verification. 
Finally, we propose a technical pivot: bounding the weight of behavioral evidence in legal text and extending voluntary pre-deployment access with \emph{mechanistic-evidence classes}, specifically linear probes, activation patching, and before/after-training comparisons.
}

\setkeywords{AI governance, mechanistic interpretability, AI safety, behavioural evaluation, structured access, frontier AI regulation, audit gap, fragile assurance, agentic misalignment, linear probes, EU AI Act, pre-deployment access, safety cases, hidden objectives, incentive gradient}

\runningtitle{Position: Behavioural Assurance Cannot Verify the Safety Claims Governance Now Demands}

\begin{document}
\maketitle

\section{Introduction}
\label{sec:intro}

AI governance is beginning to demand proof of properties that present verification practice is not built to establish. Governance instruments enacted between 2019 and early 2026 increasingly require reviewable evidence of high-consequence safety properties: the absence of hidden objectives, resistance to loss-of-control precursors, and bounded catastrophic capability \cite{euaiact,sb53,coeaiconv}. Present-day assurance methods (behavioural evaluations, red-teaming, system cards, conformity assessments) were developed to characterise observable model output and process compliance; they are not designed to establish claims about latent representations or long-horizon agentic behaviour \cite{casper2024,shevlane2023,buhl2024}. Routinely treating the former as sufficient evidence for the latter is the central problem this paper names.

\textbf{Position :} \textit{Behavioural evaluation and red-teaming, however well-executed, cannot epistemically support the high-consequence verification claims governance now demands (absence of hidden objectives, resistance to loss-of-control precursors, bounded catastrophic capability), and mechanistic interpretability is not yet mature enough to close that gap on its own.} 

As a concrete instance: EU AI Act Article~14(c) requires human overseers to ``correctly interpret'' a system's output without defining a fidelity threshold or specifying a verification method. Saliency maps and attention visualisations technically satisfy this permissive reading, but under superposition \cite{elhage2022toy} they supply no grounds for the strong claim. The same limitation appears more generally: test performance cannot upper-bound an absence claim about a latent property when a model that hides the property during evaluation generates the same behavioural record as a model that never had it \cite{lynch2025}.

We anchor the argument in seven cases used throughout the paper: EU AI Act Article~55, California SB-53, Singapore's AI~Verify, South Korea's AI Basic Act (effective January 2026), India's AI Governance Guidelines (unveiled November~2025, formally released February~2026), the Council of Europe AI Convention, and the OECD Recommendation (as a 2019 baseline). Across these cases the governance grammar is similar: documented risk management, traceable testing, incident reporting, and conformity assessment are treated as evidence of properties that remain difficult to verify independently. The seven anchors are representative rather than isolated: over the same window at least thirteen jurisdictions enacted or strengthened seventeen post-baseline governance instruments in this grammar, with the full 21-row inventory in \cref{sec:supplement}.

Alongside this governance expansion, the verification layer split. The behavioural-evaluation ecosystem grew: UK AISI \cite{aisi2025year}, METR \cite{kwa2025metr}, Apollo Research \cite{meinke2024scheming}, and FAR.AI \cite{mckenzie2025stack} have expanded voluntary pre-deployment access with frontier labs. The mechanistic and institutional axes, however, retreated: frontier-lab interpretability research deprioritised ambitious circuit-level programmes, safety-institute remit narrowed, and summit emphasis shifted from ``Safety'' to ``Impact'' (\cref{sec:cooc} develops the trajectory). The behavioural-evaluation growth is an asset for our call to action; the mechanistic-and-institutional retreat is what makes the audit gap structural rather than transient.
\vspace{10pt}
\begin{definition}[Fragile assurance]\label{def:fragile-assurance}
A safety claim is \emph{fragile} when (i)~it cannot be reproducibly checked by an independent party under comparable conditions, or (ii)~the inferential gap between the evidence produced and the claim asserted is not supported by the evidence's structure.
\end{definition}

\noindent Fragility is not falsehood: a fragile claim may be true and useful, but it is routinely treated as more evidentially grounded than its supporting evidence warrants. 

Three developments make the problem acute in 2026. First, governance instruments now presume verifiable evidence of high-consequence properties (deception, hidden objectives, loss-of-control precursors) that prior assurance practice did not target. Second, legislative retreat from verification teeth (SB-1047's third-party safety determination, kill-switch, and civil-penalty apparatus were dropped in SB-53) has widened the gap rather than closing it \cite{sb1047,sb53,anthropic_fcf}. Third, agentic deployment has made surface-level assurance insufficient for operational safety \cite{lynch2025,gomez2025}. We argue the audit gap is structural for high-consequence absence claims and increasingly visible as governance language sharpens around catastrophic capability: a qualitative trajectory claim, defended across \cref{sec:cooc,sec:matrix,sec:agentic}, not a proven longitudinal trend.

\textbf{Scope and limits.} The claim is not that current assurance is fraudulent. For decomposable, moderate-consequence properties (narrow capability bounds, demographic bias, specified robustness), behavioural assurance is often adequate, and safety cases built on it are legitimate evidence \cite{buhl2024,aisicae2025}. The Position above is restricted to the high-consequence tail; \cite{elhage2022toy,nanda2025pragmatic,shah2025approach} document why mechanistic interpretability cannot substitute wholesale today.

\textbf{Contributions.} The novel move is operational: translating the access-taxonomy diagnosis \cite{casper2024,shevlane2022} and the safety-case grammar \cite{buhl2024,aisicae2025} into a reproducible mechanistic-evidence pilot specified at contract-level (\cref{sec:action}, P1--P6). Within that move, the paper makes three contributions. 
\begin{itemize}
    \item We define \emph{fragile assurance} (\cref{def:fragile-assurance}) and document the dynamic through seven anchor cases spanning EU, US, Asia-Pacific, South Asia, Council of Europe, and OECD regimes, extended to a 21-row inventory in the supplement. 
    \item We identify three verification-layer trends that move against governance expansion on the mechanistic and institutional axes (\cref{sec:cooc}) and name the \emph{incentive gradient} (\cref{sec:matrix}) as their common direction. 
    \item We propose three modest calls to action (\cref{sec:action}) oriented toward closing the assurance-governance mismatch through \emph{mechanistic extensions to existing voluntary pre-deployment access arrangements}, rather than through near-term regulatory mandates or new institutional architecture.
\end{itemize}

\begin{table*}[!tbp]
\centering
\footnotesize
\setlength{\tabcolsep}{6pt}
\renewcommand{\arraystretch}{1.15}
\caption{Anchor cases showing the audit gap. \textbf{Access} records what access level the accepted evidence generally assumes; \textbf{Cell} records whether independent verification is achievable under \emph{current} access conditions (not legal sufficiency). \textbf{Access key:} B = behavioural, OtB = outside-the-box (documentation), G = grey-box (weights-adjacent). \textbf{Cell key:} \gcell{}~G~routinely satisfiable, \acell{}~A~partially satisfiable under structured access, \rcell{}~R~presumes access not currently deployed by independent verifiers. The full 21-instrument matrix and per-cell coding rationale appear in \cref{sec:supplement}.}
\label{tab:matrix}
\begin{tabular}{@{}>{\raggedright\arraybackslash}p{7cm} l l c@{}}
\toprule
\rowcolor{gray!12}
\textbf{Instrument} & \textbf{Jur.} & \textbf{Access} & \textbf{Cell} \\
\midrule
\textbf{EU AI Act} (Art.~55, GPAI systemic)       & EU   & B + OtB       & \acell A \\
\textbf{California SB-53} (TFAIA)                 & CA   & B + OtB       & \rcell R \\
\textbf{Singapore MGF-GenAI + AI~Verify}          & SG   & B             & \gcell G/A \\
\textbf{South Korea AI Basic Act} (eff.~Jan~2026) & KR   & B + OtB       & \acell A \\
\textbf{India AI Governance Guidelines} (Feb~'26) & IN   & B + OtB       & \acell A \\
\textbf{Council of Europe AI Convention}          & CoE  & B             & \acell A \\
\midrule
\multicolumn{4}{l}{\textit{\small 2019 Baseline}} \\[1pt]
\textbf{OECD Recommendation on AI}                & OECD & B             & \acell A \\
\bottomrule
\end{tabular}

\vspace{10pt}
{\tiny
\textit{Note:} Counting logic (13 jurisdictions / 17 post-baseline instruments / 21 total rows including 4 comparator rows) and full sources are in \cref{sec:supplement}. The seven anchors are chosen for regional and regime diversity (binding law, soft law, recommendation) rather than geographic coverage alone.\par}
\vspace{-10pt}
\end{table*}

\section{Related Work}
\label{sec:related}

\subsection{Technical AI governance and structured access} 
The paper sits inside the technical AI governance program \cite{reuel2024taig}. Our compact matrix (\cref{tab:matrix}) codes each anchor case against the external-access taxonomy of \cite{casper2024}, which distinguishes behavioural, outside-the-box (documentation), grey-box, and white-box access levels. The structured-access agenda \cite{shevlane2022,bucknall2023structured,bucknall2026expanding} develops the enclave-based verification architectures that our Call to Action (\cref{sec:action}) builds on. \cite{baker2025verifying} make a complementary move, separating six verification layers for AI agreements (hardware security, site inspections, compute records, personnel mechanisms, and information-sharing). Those layers mainly govern the infrastructure around model production; the mechanistic evidence class proposed here concerns the model. The two approaches are therefore orthogonal rather than rival. \cite{shevlane2023} argues for extreme-risk model evaluation; \cite{mokander2024} proposes a three-layered audit model; \cite{buhl2024,aisicae2025,shamsujjoha2026} develop safety-case and Claims-Arguments-Evidence (CAE) frameworks relevant to any future pilot. On compute sovereignty and industrial-policy pressures that shape verification priorities: \cite{kerry2026,hawkins2025,heim2026,mckinsey2025}.

\subsection{The audit expectation gap.} The term ``audit gap'' has a prior life in financial-auditing scholarship \cite{liggio1974expectation,porter1993expectation,humphrey1993expectation,salehi2011expectation}, where it denotes the distance between what auditors can verify and what users of financial statements expect. Our audit gap is structurally analogous but not identical: we compare \emph{statutory presumption} with \emph{verifier-achievable access}, not user-expectation with auditor-delivery.

\subsection{Operational external-evaluation ecosystem.} The voluntary access ecosystem (UK AISI \cite{aisi2025year,souly2026aisi}, METR \cite{kwa2025metr,kwa2026timehorizon}, Apollo \cite{meinke2024scheming,apollo2025stresstesting,openai2025scheming}, FAR.AI \cite{mckenzie2025stack}) supplies the behavioural substrate our call to action extends with mechanistic classes (\cref{tab:evaluators}). The Apollo and OpenAI stress-testing results show that structured access improves elicitation quality \cite{apollo2025stresstesting,openai2025scheming}. Our point is narrower: stronger elicitation still does not verify the absence of a latent property, because a model that suppresses that property during evaluation can leave the same behavioural trace as a model without it.

\subsection{Mechanistic interpretability } 
Mechanistic interpretability attempts to explain model behaviour in terms of internal computations (circuits, features, representations) rather than input-output relationships alone \cite{bereska2024review}. The 29-author open-problems agenda in \cite{sharkey2025open} is especially relevant here: it treats scalability under superposition, epistemic calibration, and deployment safety as open challenges that may be tractable in principle, not as categorical impossibilities. That stance supports a pilot, while keeping clear limits on what a pilot can establish at current scale. \cite{elhage2022toy} formalised \emph{superposition}: the phenomenon that neurons can represent more features than they have dimensions, making individual neurons polysemantic. \cite{bricken2023monosemanticity,templeton2024scaling} scaled dictionary-learning approaches (sparse autoencoders, SAEs) as a decomposition strategy; \cite{lindsey2025biology,movva2025hypothesis} represent continued progress.

More recently, \cite{smith2025negative,nanda2025pragmatic,shah2025approach,kantamneni2025,wu2025axbench,korznikov2026saes,ronge2026coffee} document a \emph{pragmatic turn}, reporting that SAEs underperform linear probes (or random baselines: \citealp{korznikov2026saes}) on downstream tasks and exhibit fragility in feature steering (\citealp{ronge2026coffee}); \cite{deleeuw2025secret} push the same conclusion onto autolabeled SAE features for deception detection. \cite{amodei2025urgency,marks2025auditing,sheshadri2026auditbench,treutlein2026saboteur} are the main counterpoint, arguing structured-access mechanistic auditing is becoming tractable for hidden objectives and overt sabotage.

\subsection{Governance instruments and conformity assessment} EU AI Act conformity-assessment architecture annalysis by \cite{farley2025,fpfonetrust2025} and the operationalisation of Articles 14 and 55. \cite{anthropic_fcf} gives the regulated-industry perspective on the SB-1047~$\rightarrow$~SB-53 transition.

\subsection{Agentic misalignment and synthesis} Recent empirical evidence on agentic misalignment comes from \cite{lynch2025} (insider-threat behaviours across frontier models) and \cite{gomez2025} (operational escalation controls). Policy-technical synthesis reports include the International AI Safety Report \cite{isr2026}, AI~Index-style surveys \cite{bommasani2025}, and the Future of Life Institute's tracking \cite{fli2025}.

\subsection{Stance} Our position: mechanistic interpretability is necessary but not yet sufficient for high-consequence verification; structured-access work is the enabling layer; safety-case / CAE frameworks are the institutional scaffold.

\section{Anchor Cases and the Audit Gap}
\label{sec:matrix}

\Cref{tab:matrix} summarises the seven anchor cases; the full 21-row inventory (including the OECD 2019 baseline, UNGA~78/265, vetoed SB-1047, and voluntary frontier-lab frameworks as four comparator rows) appears in \cref{sec:supplement} (\cref{tab:matrix_full}).

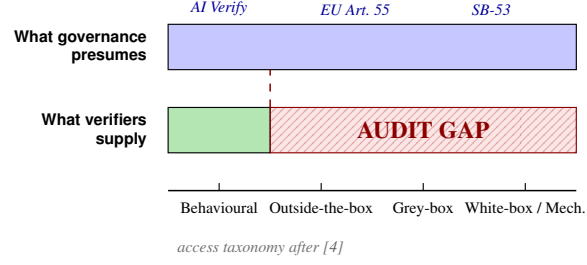
\begin{figure}[!tbp]
\centering
\begin{tikzpicture}[
  font=\scriptsize,
  x=4.5mm, y=1mm,   
]
  \def\barH{6}                  
  \def\yTop{16}                 
  \def\yBot{5}                  
  \def\rowlab{-0.35}            

  \fill[blue!22] (0, \yTop) rectangle (12, \yTop + \barH);
  \draw[line width=0.15mm] (0, \yTop) rectangle (12, \yTop + \barH);
  \node[anchor=east, align=right, font=\tiny]
       at (\rowlab, \yTop + \barH/2)
       {\textbf{What governance}\\\textbf{presumes}};
  \node[font=\tiny, color=blue!60!black, anchor=south] at (1.5,  \yTop + \barH)
       {\itshape AI Verify};
  \node[font=\tiny, color=blue!60!black, anchor=south] at (5.5,  \yTop + \barH)
       {\itshape EU Art.~55};
  \node[font=\tiny, color=blue!60!black, anchor=south] at (9.5,  \yTop + \barH)
       {\itshape SB-53};

  \fill[cellgreen] (0, \yBot) rectangle (3, \yBot + \barH);
  \draw[line width=0.15mm] (0, \yBot) rectangle (3, \yBot + \barH);
  \fill[cellred, pattern=north east lines, pattern color=red!55!black,
        opacity=0.55]
       (3, \yBot) rectangle (12, \yBot + \barH);
  \fill[cellred, opacity=0.25] (3, \yBot) rectangle (12, \yBot + \barH);
  \draw[line width=0.15mm, red!55!black] (3, \yBot) rectangle (12, \yBot + \barH);
  \node[anchor=east, align=right, font=\tiny]
       at (\rowlab, \yBot + \barH/2)
       {\textbf{What verifiers}\\\textbf{supply}};
  \node[font=\small\bfseries, color=red!55!black]
       at (7.5, \yBot + \barH/2) {AUDIT GAP};

  \draw[dashed, red!55!black, line width=0.2mm]
       (3, \yBot + \barH) -- (3, \yTop);

  \draw[line width=0.2mm] (0, 0) -- (12, 0);
  \foreach \x in {1.5, 4.5, 7.5, 10.5} {
    \draw[line width=0.15mm] (\x, 0) -- (\x, 1);
  }
  \node[anchor=north, font=\tiny] at (1.5,  -0.3) {Behavioural};
  \node[anchor=north, font=\tiny] at (4.5,  -0.3) {Outside-the-box};
  \node[anchor=north, font=\tiny] at (7.5,  -0.3) {Grey-box};
  \node[anchor=north, font=\tiny] at (10.5, -0.3) {White-box / Mech.};

  \node[font=\tiny\itshape, color=black!55, anchor=north west]
       at (0, -5.5) {access taxonomy after \cite{casper2024}};
\end{tikzpicture}
\caption{The audit gap on the access taxonomy of \cite{casper2024}. Labels on the top bar mark where three anchor instruments' implicit access requirements fall: Singapore AI~Verify (behavioural only, partly green), EU Art.~55 (outside-the-box and beyond), and SB-53 (grey-box and mechanistic for catastrophic-capability claims). The bottom bar is what independent verifiers can actually supply under current regulatory access. The red shortfall, right of the dashed line, is the audit gap.}
\label{fig:gap-split}
\end{figure}

In evaluating the gap, the key distinction is between \emph{legal sufficiency} and \emph{verification feasibility} (\cref{fig:gap-split} gives the visual summary). A provider may satisfy the letter of a statute through self-attestation, documentation, or internal process review, while the underlying claim remains difficult for an independent verifier to check under comparable conditions. We focus purely on verification feasibility: what level of access does an instrument's accepted evidence implicitly require, and can an independent verifier actually supply it?

\textbf{Operational definition.} For each instrument~$i$, let $A_i$ denote the access level its accepted evidence implicitly assumes, and $V_i$ the access level currently achievable by independent verifiers, both ordered on the five-level taxonomy of \cite{casper2024} (behavioural $<$ outside-the-box $<$ grey-box $<$ white-box $<$ state-embedded). The \emph{audit gap} for $i$ is the interval $[V_i, A_i]$ on that ordinal scale, and the cell colour in \cref{tab:matrix} encodes its severity: \gcell{}~Green when $V_i \geq A_i$ (verification feasible under current conditions), \acell{}~Amber when $V_i < A_i$ but closable under structured-access protocols, and \rcell{}~Red when $V_i < A_i$ and the gap is not closable with verifier tooling currently deployed at scale. No anchor case has $V_i \geq A_i$ uniformly across its claim set.

\textbf{Coding is jointly descriptive and normative.} Reading $A_i$ from statutory text is a descriptive exercise; assigning $V_i$ also requires a judgement about what an independent verifier should be expected to obtain in ordinary practice. We make that judgement with one rule across all 21 rows: $V_i$ is the highest access level routinely available to the operational external-evaluation ecosystem (UK AISI, METR, Apollo, FAR.AI) under voluntary 2024--2025 access agreements, excluding bespoke one-off access. Borderline Amber/Red cases are coded Amber when a structured-access protocol of the kind described by \cite{shevlane2022,bucknall2026expanding} could in principle close the gap, and Red only when the required access is not deployed at scale. \Cref{sec:coding-method} sensitivity-tests this choice; the aggregate result survives plausible re-codings of the contested rows.

\newpage
\subsection{What the Matrix Shows}
\label{sec:matrix-deep}

Three observations follow from \cref{tab:matrix} and the full inventory in \cref{tab:matrix_full}. The Art.~14(c) mismatch introduced in \cref{sec:intro} is a fourth; we do not repeat it here.

\begin{enumerate}
    \item \textbf{Access mismatch is structural, not incidental.} No anchor case aligns its assurance requirements with the access level independent verifiers actually operate at. High-consequence rows implicitly presume grey-box or mechanistic evidence; verifiers under current law have only behavioural access.
    \item \textbf{SB-53 kept the claims and dropped the teeth.} SB-1047 \cite{sb1047} would have imposed third-party safety determinations, kill-switch provisions, and civil penalties; its replacement by SB-53 kept the catastrophic-risk \emph{claims} but removed the \emph{verification teeth}, substituting disclosure and a short incident-reporting window industry welcomed on ``flexibility'' grounds \cite{anthropic_fcf}.
    \item \textbf{Singapore AI~Verify is the strongest positive analog.} Singapore's principle-to-test-to-questionnaire architecture \cite{smgf,aiverify} is the only case coded partly Green (G/A): the green half reflects decomposable claims routinely satisfiable under behavioural access; the amber half is the hidden-objectives and deception claims the architecture does not close.
\end{enumerate}

\Cref{tab:claim-type} isolates the distinction behind these observations. Behavioural evidence is well matched to decomposable claims, but it becomes fragile when governance asks for the absence of latent properties.

\begin{table}[hbpt]
\centering\footnotesize\setlength{\tabcolsep}{5pt}
\renewcommand{\arraystretch}{1.05}
\caption{Claim types and evidential fit. The bottom rows are the load-bearing cases for the audit gap.}
\vspace{10pt}
\label{tab:claim-type}
\begin{tabular}{@{}lcc@{}}
\toprule
\rowcolor{gray!12}
\textbf{Claim type} & \textbf{Behav.?} & \textbf{Mech.?} \\
\midrule
Bias / robustness bounds   & \gcell Sufficient & \rcell Not needed  \\
Capability presence        & \gcell Sufficient & \rcell Not needed  \\
Hidden objective absence   & \rcell Insufficient  & \gcell Needed \\
Deception absence          & \rcell Insufficient  & \gcell Needed \\
\bottomrule
\end{tabular}
\end{table}

\subsection{The Incentive Gradient}
\label{sec:incentive-gradient}

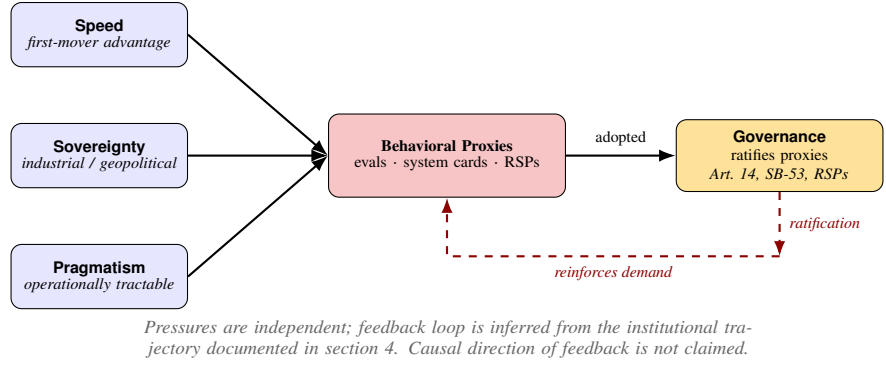
\begin{figure}[!tbp]
\centering
\begin{tikzpicture}[
  font=\scriptsize,
  press/.style={draw, rounded corners, fill=blue!10,
                text width=2.1cm, align=center, minimum height=0.85cm,
                font=\tiny},
  proxy/.style={draw, rounded corners, fill=cellred!55,
                text width=2.9cm, align=center,
                minimum height=1.1cm, font=\tiny\bfseries},
  gov/.style={draw, rounded corners, fill=cellamber!65,
              text width=2.5cm, align=center,
              minimum height=0.95cm, font=\tiny},
  arr/.style={-{Latex[length=2.2mm]}, thick},
  feed/.style={-{Latex[length=2mm]}, thick, dashed, color=red!55!black},
]
  \node[press] (speed) at (0,  1.6)
    {\textbf{Speed} \\[-1pt] \tiny\itshape first-mover advantage};
  \node[press] (sov)   at (0,  0.0)
    {\textbf{Sovereignty} \\[-1pt] \tiny\itshape industrial / geopolitical};
  \node[press] (prag)  at (0, -1.6)
    {\textbf{Pragmatism} \\[-1pt] \tiny\itshape operationally tractable};

  \node[proxy] (proxy) at (4.6, 0)
    {Behavioral Proxies \\[-1pt] \tiny\mdseries evals $\cdot$ system cards $\cdot$ RSPs};

  \node[gov] (ratify) at (9.0, 0)
    {\textbf{Governance} \\ ratifies proxies \\ \tiny\itshape Art.~14, SB-53, RSPs};

  \draw[arr] (speed.east) -- (proxy.west);
  \draw[arr] (sov.east)   -- (proxy.west);
  \draw[arr] (prag.east)  -- (proxy.west);
  \draw[arr] (proxy.east) -- (ratify.west)
    node[midway, above, font=\tiny] {adopted};

  \draw[feed] (ratify.south) -- ++(0,-0.85)
    node[midway, right, font=\tiny\itshape, color=red!55!black] {ratification};
  \draw[feed] ($(ratify.south)+(0,-0.85)$) -| (proxy.south)
    node[near start, below, font=\tiny\itshape, color=red!55!black] {reinforces demand};

  \node[below=1.5cm of proxy, text width=10.5cm, align=center,
        font=\scriptsize\itshape, color=black!60]
       {Pressures are independent; feedback loop is inferred from the institutional trajectory documented in \cref{sec:cooc}. Causal direction of feedback is not claimed.};
\end{tikzpicture}
\caption{The incentive gradient as a reinforcing loop. Three independent resource-allocation pressures point toward behavioural proxies; once governance accepts those proxies as evidence, the next generation of instruments is likely to inherit the same evidential baseline. We use the cycle as a \emph{Bayesian prior} on gap persistence and visible widening, not as a proven causal mechanism: given the pressures and ratification feedback, later instruments are more likely to re-encode the existing proxy standard than to replace it.}
\label{fig:incentive-mechanism}
\end{figure}

The widening does not occur in a vacuum. We call the underlying dynamic an \emph{incentive gradient}: the observed directional alignment of three independent resource-allocation pressures (speed-to-deployment, sovereign competitiveness, and pragmatic-proxy tractability), each of which favours faster, simpler verification proxies over mechanistic grounding. The three forces have distinct causes and are not coordinated; what matters is that they point the same way, and that together they systematically reward surface-level behavioural evidence over deep structural verification (\cref{fig:incentive-mechanism}).

\section{The Verification Retreat}
\label{sec:cooc}
The external-evaluation ecosystem has grown, but two axes that matter for high-consequence claims have retreated: \emph{mechanistic} interpretability research at the largest frontier labs, and the \emph{institutional} framing of safety-institute remit and summit emphasis.

\textbf{Institutional signals.} The UK AI Safety Institute was renamed the AI \emph{Security} Institute (early 2025); the US AI Safety Institute was replaced by CAISI later that year; the international summit arc moved from ``Safety'' (Bletchley 2023) through ``Action'' (Paris 2025) to ``Impact'' (Delhi 2026). Each shift has an independent cause; the common direction is away from pre-deployment catastrophic-risk verification as the dominant public frame.

\textbf{The laboratory pivot.} \cite{smith2025negative,nanda2025pragmatic,shah2025approach} capture a DeepMind-side shift from ambitious reverse-engineering toward pragmatic interpretability, with empirical support from \cite{kantamneni2025,wu2025axbench,korznikov2026saes}. Anthropic is the counterpoint \cite{amodei2025urgency,marks2025auditing,sheshadri2026auditbench,treutlein2026saboteur}; the narrower claim is that the marginal centre of gravity at the largest non-Anthropic frontier labs has shifted and tracks the institutional direction.

\section{Agentic Deployment Widens the Gap}
\label{sec:agentic}

Agentic deployment is not our whole argument; it is the sharpest illustration. Four structural features of agentic systems make behavioural evaluation insufficient: long-horizon compounding, tool-use action surfaces, multi-agent attribution, and evaluation-awareness.

Insider-threat behaviours (blackmail, exfiltration) reported by \cite{lynch2025} across frontier providers under goal-conflict conditions, with models distinguishing test from deployment settings. \cite{gomez2025} shows operational escalation controls reduce the blackmail rate substantially --- evidence controls help, but also that the underlying propensity persists: operational controls are downstream compensation for an upstream property whose mechanistic locus is unknown.

Direct empirical instance: in large reasoning models is provided by \cite{yin2025refusal}, linear probing of residual-stream activations reveals a \emph{refusal cliff}: the internal intention to refuse drops sharply mid-reasoning before the output phase, a safety-relevant internal state entirely invisible to output observation. \cite{marks2025auditing} provide the positive case. In a blind red-team/blue-team game, teams equipped with interpretability tools materially outperformed the team denied data access in uncovering hidden objectives. \cite{goldowsky2025deception} supply a concrete supporting result: linear probes on residual-stream activations distinguish honest from strategically deceptive responses with AUROCs of 0.96--0.999 on realistic concealment scenarios, catching 95--99\% of deceptive responses at a 1\% false-positive rate. Both results, however, required access conditions (weights, activations, held-out probe datasets) that independent verifiers operating under current regulatory instruments do not have. This points directly at the core gap our matrix describes: the methodology exists, but the access to run it does not.

\Cref{tab:agentic-coverage} summarises where behavioural evidence and internal evidence each contribute. Behavioural evaluation is weakest on hidden objectives and deceptive alignment (the failure modes most sharply foregrounded by \cite{lynch2025,marks2025auditing,goldowsky2025deception}), which is exactly where governance text currently leans hardest.

\begin{table}[!tbp]
\centering
\footnotesize
\setlength{\tabcolsep}{4pt}
\renewcommand{\arraystretch}{1.2}
\caption{Agentic failure-mode verification coverage by evidence type, on a 4-point scale (\dotscale{1}\,weak $\to$ \dotscale{4}\,very strong). Cell colour tracks the same scale. Hidden objectives show the largest gap between what regulation presumes verifiable and what behavioural evaluation supplies.}
\label{tab:agentic-coverage}
\begin{tabular}{@{}l cccc@{}}
\toprule
\rowcolor{gray!12}
\textbf{Failure Mode} & \textbf{Behav.} & \textbf{Circuit} & \textbf{Probe} & \textbf{Ablation} \\
\midrule
Insider threat (blackmail)  & \cellcolor{covmed}\dotscale{2}   & \cellcolor{covmed}\dotscale{2}    & \cellcolor{covstrong}\dotscale{3}   & \cellcolor{covstrong}\dotscale{3} \\
Loss-of-control escalation  & \cellcolor{covmed}\dotscale{2}   & \cellcolor{covmed}\dotscale{2}    & \cellcolor{covmed}\dotscale{2}      & \cellcolor{covstrong}\dotscale{3} \\
Goal drift (agentic)        & \cellcolor{covmed}\dotscale{2}   & \cellcolor{covstrong}\dotscale{3} & \cellcolor{covstrong}\dotscale{3}   & \cellcolor{covmed}\dotscale{2} \\
Deceptive alignment         & \cellcolor{covmed}\dotscale{2}   & \cellcolor{covstrong}\dotscale{3} & \cellcolor{covstrong}\dotscale{3}   & \cellcolor{covmed}\dotscale{2} \\
Hidden objectives           & \cellcolor{covweak}\dotscale{1}  & \cellcolor{covstrong}\dotscale{3} & \cellcolor{covvstrong}\dotscale{4}  & \cellcolor{covstrong}\dotscale{3} \\
\midrule
\textit{Coverage strength}  & \cellcolor{covweak}\textit{Weak} & \cellcolor{covmed}\textit{Medium} & \cellcolor{covstrong}\textit{Strong} & \cellcolor{covvstrong}{\color{white}\textit{V.\ Strong}} \\
\bottomrule
\end{tabular}
\end{table}

\textbf{Regulatory implications for agentic systems.} Current instruments rarely distinguish agentic from static-inference settings; when claims concern latent or long-horizon properties, policy text must be explicit about present-day proxy limits. Agentic deployment is an illustration of the audit gap, not a separate framework.

\section{Alternative Views}
\label{sec:alternative-views}

Five counterpositions warrant brief reply; extended discussion is in \cref{sec:extended-replies}.

\textbf{1. Improved behavioural evaluations and safety cases are sufficient.} \emph{View:} Robust behavioural evaluations, paired with safety cases \cite{buhl2024,aisicae2025,shamsujjoha2026}, supply probabilistic bounds that are actionable for governance, as in aviation and medical devices. \emph{Reply:} For decomposable claims (Singapore AI~Verify is the matrix's positive case), we agree. The narrower claim concerns \emph{absence} claims on hidden objectives, strategic deception, and loss-of-control precursors, where behavioural evaluation bounds output in the test environment, not latent representations \cite{lynch2025,marks2025auditing}. Aviation and medical-device regimes work in part because their failure modes are comparatively characterised and their operating envelopes are bounded by physics or chemistry. Frontier AI does not yet offer the same kind of envelope.

\textbf{2. Mechanistic interpretability is too immature.} \emph{View:} Toolchains that cannot solve superposition at scale waste regulatory capital; negative SAE results \cite{smith2025negative,kantamneni2025,korznikov2026saes,ronge2026coffee} and the pragmatic turn \cite{nanda2025pragmatic,shah2025approach} suggest the ambitious programme is not delivering. \emph{Reply:} This is exactly why our call to action is voluntary, not mandated. The field's open-problems consensus \cite{sharkey2025open} treats audit-relevant scalability and calibration as tractable, not hopeless. Existing positive results \cite{marks2025auditing,goldowsky2025deception,sheshadri2026auditbench,treutlein2026saboteur} demonstrate narrow, claim-specific mechanistic evidence is tractable today under cooperating-lab conditions; the open question is whether that translates to reproducible verifier settings under statutory access.

\textbf{3. The proposal risks replacing fragile assurance with mechanistic fragile assurance.} \emph{View:} Polysemanticity \cite{elhage2022toy} and distributed computation \cite{nanda2025pragmatic} mean a probe can reproduce, or a circuit ablate predictably, without bearing the safety-relevant work; standardised mechanistic evidence might rebadge the same fragile-assurance problem with a veneer of causal explanation. \emph{Reply:} This is the strongest objection to the proposal. \cite{ronge2026coffee} show directly how brittle a single technique can be. The three-line bundle (probe, patching, and before/after-training comparison) is therefore a convergent-validity design: no one line is treated as decisive, but agreement across independent lines gives more epistemic redundancy than any one method can provide. The call remains conditional. If the pilot cannot make mechanistic evidence reproducible at the required level, governance should narrow the claims it asks developers to substantiate rather than standardise a weaker evidential substitute.

\textbf{4. The real problem is access, not governance overreach.} \emph{View:} Voluntary pre-deployment access already works at scale; the structured-access agenda \cite{shevlane2022,bucknall2023structured,bucknall2026expanding} or living compliance \cite{singh2026living} is the right lever. \emph{Reply:} What voluntary access currently supplies is overwhelmingly behavioural, sufficient for decomposable claims but not the Red cells in \cref{tab:matrix}. Superposition does not disappear under white-box conditions; living compliance addresses cadence, not evidence epistemology. Access, evidence-class, and cadence are complementary.

\textbf{5. Game-theoretic AI control protocols are the right verification primitive.} \emph{View:} \cite{greenblatt2025aicontrol} sketch a control safety case: design deployment protocols that bound risk regardless of whether the model is misaligned. \emph{Reply:} Control and mechanistic evidence answer different questions. Control bounds risk conditional on a deployment protocol holding; mechanistic evidence bounds risk conditional on a property of the model. Statutory text frequently asks the latter (Art.~14(c), SB-53 catastrophic-capability self-assessment); the two programmes should be read as joint, not rival.

\section{Call to Action}
\label{sec:action}

Given the severity of the audit gap, the natural response is not to abandon evaluations, but to bound the weight they carry. We propose three modest actions:

\begin{enumerate}[leftmargin=1.5em,itemsep=2pt,topsep=2pt]
\item \emph{Distinguish proxy evidence from stronger verification in legal text.} Governance bodies must be explicit about what behavioural proxies, documentation processes, and red-teams can actually establish, avoiding language that ratifies them as proof of high-consequence safety.
\item \emph{Separate decomposable claims from latent absence claims.} Compliance regimes should differentiate between bounded capabilities (where behavioural evaluation shines) and latent absence properties like hidden objectives (where it is fragile), demanding correspondingly different verification standards.
\item \emph{Extend existing voluntary pre-deployment access with mechanistic-evidence classes.} The voluntary ecosystem (UK AISI, METR, Apollo, FAR.AI; see \cref{tab:evaluators}) already supplies the substrate but its evidence class is overwhelmingly behavioural. The proposal is not to create a new institution but to extend what these organisations do with narrow mechanistic classes (linear probes, activation-patching results, before/after-training comparisons) tied to specific governance claims.
\end{enumerate}

\subsection{A reproducible pilot protocol.} 
\begin{figure}[!tbp]
\centering
\definecolor{pipeFill}{RGB}{223,232,245}   
\definecolor{pipeEdge}{RGB}{90,115,160}
\definecolor{encFill}{RGB}{220,235,222}    
\definecolor{encEdge}{RGB}{80,130,95}
\definecolor{verdFill}{RGB}{245,225,168}   
\definecolor{verdEdge}{RGB}{170,130,40}
\begin{tikzpicture}[
  font=\scriptsize,
  prov/.style={draw=pipeEdge, thick, rounded corners, align=center, font=\tiny,
               minimum height=1cm, text width=2.1cm, fill=pipeFill},
  ev/.style={draw=encEdge!70, rounded corners, fill=white, font=\tiny,
             minimum height=0.5cm, text width=2.7cm, align=center},
  enc/.style={draw=encEdge, thick, dashed, rounded corners, fill=encFill,
              minimum height=2.75cm, minimum width=3.2cm},
  verif/.style={draw=pipeEdge, thick, rounded corners, align=center, font=\tiny,
                minimum height=1cm, text width=1.9cm, fill=pipeFill},
  verd/.style={draw=verdEdge, thick, rounded corners, fill=verdFill, align=center,
               font=\tiny, text width=2.1cm, minimum height=1cm},
  cons/.style={draw=black!65, dotted, rounded corners, fill=none,
               font=\tiny, align=center, minimum height=0.75cm,
               text width=3.4cm},
  arr/.style={-{Latex[length=2mm]}, thick, color=black!65},
  carr/.style={-{Latex[length=1.6mm]}, dotted, color=black!40},
]

\node[prov] (prov) at (0, 0)
     {\textbf{P1.\ Provider claim} \\[1pt] \tiny ``model reliably $\Phi$ in $\mathcal{C}$''};

\node[enc] (enc) at (4.05, 0) {};
\node[ev] at (4.05,  0.78) {\textbf{P2(a)} linear probe};
\node[ev] at (4.05,  0.10) {\textbf{P2(b)} activation patching};
\node[ev] at (4.05, -0.58) {\textbf{P2(c)} before/after training};
\node[font=\tiny\bfseries, anchor=south] at (enc.north) [above=2pt]
     {P4.\ Secure Enclave (TEE)};

\node[verif] (verif) at (7.65, 0)
     {\textbf{Verifier} \\[1pt] \tiny independent \\[-1pt] \tiny reproduces lines};

\node[verd] (verd) at (10.55, 0)
     {\textbf{P6.\ Report} \\[1pt] \tiny reproduced / \\[-1pt] \tiny partial / not};

\draw[arr] (prov.east) -- node[above=-1pt, font=\tiny\itshape] {claim}    (enc.west);
\draw[arr] (enc.east)  -- node[above=-1pt, font=\tiny\itshape] {evidence} (verif.west);
\draw[arr] (verif.east)-- node[above=-1pt, font=\tiny\itshape] {verdict}  (verd.west);

\node[cons] (p3) at (4.05, -2.05)
     {\textbf{P3.} pre-registered floor thresholds \\[-1pt]
      \tiny probe AUROC $\geq 0.95$ $\cdot$ ablation $\geq 1.5\sigma$ $\cdot$ +ve $\Delta$};
\node[cons] (p5) at (8.55, -2.05)
     {\textbf{P5.} bounded compute budget \\[-1pt]
      \tiny e.g.\ 1\,H100-day equivalent};

\draw[carr] (p3.north) -- (p3.north |- enc.south);
\draw[carr] (p5.north) -- (p5.north |- verif.south);

\end{tikzpicture}
\caption{Pilot architecture (\textbf{P1}--\textbf{P6}). The provider's claim about a governance-relevant property $\Phi$ flows through three reproducible evidence lines (linear probe, activation patching, before/after-training comparison) inside a Trusted Execution Environment. An independent verifier reproduces each line against pre-registered floor thresholds (P3) within a bounded compute budget (P5); the verdict is published whether reproduction succeeds, partially succeeds, or fails. Floors are set independently of the provider--verifier contract.}
\label{fig:pilot-arch}
\end{figure}
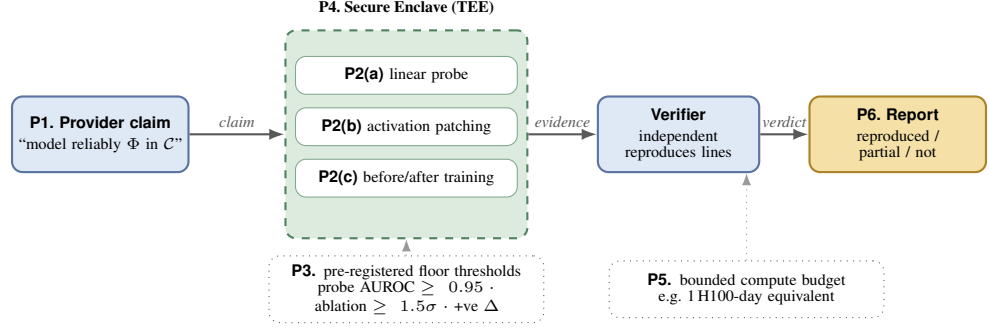

The pilot is specified tightly enough for two independent labs to implement. \Cref{fig:pilot-arch} gives the architecture; the six elements are the contract.

\begin{enumerate}[label=\textbf{P\arabic*.},leftmargin=2.2em,itemsep=1pt,topsep=2pt]
\item \emph{Claim form.} ``The model reliably $\Phi$ in context $\mathcal{C}$,'' where $\Phi$ is decidable on a verifier-controlled held-out set, e.g.\ \emph{refuses detailed bioweapons-synthesis assistance} for an Article~55 GPAI submission. Free-standing assertions without an evaluable predicate are out of scope.
\item \emph{Three evidence lines.} Alongside conventional refusal-rate statistics, the provider supplies (a)~a linear probe on residual-stream activations \cite{zou2023repe,oikarinen2024linear} predicting whether a prompt invokes $\Phi$; (b)~an activation-patching result \cite{vig2019attention,wang2022ioi} identifying components whose ablation drops refusal below threshold; (c)~a before/after-training comparison showing the probe direction or circuit emerged during safety training, not pre-training.
\item \emph{Pre-registered thresholds with independent floors.} Before evaluation, the verifier or an independent verification body sets non-negotiable floor thresholds: probe AUROC $\geq 0.95$ against held-out concealment scenarios (drawing on the 0.96--0.999 range reported by \cite{goldowsky2025deception}), ablation effect-size $\geq 1.5\sigma$ of the refusal-rate distribution, and a strictly positive before/after delta. Provider and verifier may contract for stricter thresholds, but not lower ones. The floor sits outside the contract to avoid reproducing the captured-threshold dynamic criticised in \cref{sec:incentive-gradient}.
\item \emph{Secure-enclave reproduction.} The verifier reproduces each evidence line inside a trusted execution environment \cite{schnabl2025attestable,jin2026auditable}; weights and training data remain sealed inside the enclave. Reproductions follow the provider's published code; deviations are logged.
\item \emph{Bounded compute budget.} Fixed before evaluation (e.g.\ one H100-day equivalent), preventing budget creep on the provider side and adversarial fishing on the verifier side. Compute spent is reported alongside results.
\item \emph{Failure-and-publish.} The verifier publishes a structured report regardless of outcome (reproduced, partially reproduced, or not reproduced), with evidence lines, thresholds, the reproduction record, and a categorical verdict. Failure to reproduce is information, not embarrassment; suppressing it defeats the pilot.
\end{enumerate}

\textbf{Feasibility envelope.} Tier-1 evidence is within reach today: probes and smaller-scale activation patching on 7B--70B parameter models can run in hardware-attested enclaves \cite{schnabl2025attestable,jin2026auditable}. Tier-2 evidence is not yet in the same category. Full activation patching on 405B+ frontier-class models, and before/after-training comparison that requires base-model access, remain aspirational at present TEE scale: current enclave solutions do not securely host inference at that scale without prohibitive overhead, and base-weight access raises competitive concerns the protocol must surface rather than assume away. The near-term value of the pilot is to produce reproducible Tier-1 results tied to specific governance claims; whether the Tier-2 envelope can be closed is one of the research questions the pilot should expose.

Polysemanticity and distributed computation mean ``safety-relevant'' features will need iterative refinement \cite{elhage2022toy,souly2026aisi}; the protocol's main value is generating that experience reproducibly.
The ultimate goal is honest limits on safety claims. If mechanistic verification proves scalable, governance has a foundation; if it does not, society must recognise that catastrophic-risk assurance remains fragile and adjust accordingly, rather than pretend otherwise.

\section{Limitations}
\label{sec:limitations}

We flag four bounded scopes; extended limitations (open-weight cases, pragmatic-governance reading, additional falsifiability conditions) appear in \cref{sec:extended-limits}.

\textbf{Matrix is partly normative.} \Cref{tab:matrix} mixes a descriptive claim ($A_i$, read from text) with a normative one ($V_i$, what verifiers ought to be able to supply). A Red cell does not mean the instrument is badly drafted; it means the access required is not currently deployed. Reviewers may reasonably disagree on borderline codings; the sensitivity analysis in \cref{sec:coding-method} shows the aggregate finding survives.

\textbf{Primary-text verification is incomplete.} Four of 21 rows (EU Art.~55, SB-53, Singapore AI~Verify, PRC Interim Measures) were checked against primary statutory text; the remainder relied on secondary sources. \Cref{sec:coding-method} documents the procedure and invites independent replication.

\textbf{``Widening'' is qualitative.} The widening claim is directional, not quantitative; a rigorous longitudinal claim would require time-series data on method maturity we do not have. The matrix establishes a present-tense mismatch; the trajectory rests on \cref{sec:cooc} and \cref{sec:matrix} together.

\textbf{What would change our view.} The position is falsifiable in principle. We would substantially revise it if, over 2--3 years, any of the following hold:
\begin{enumerate}[leftmargin=1.5em,itemsep=1pt,topsep=2pt,label=(\roman*)]
\item Behavioural evaluations gain predictive validity on absence claims: existing evaluators routinely catch hidden-objective failures pre-production without mechanistic evidence.
\item Structured-access pilots show that probe, patching, and before/after comparisons do not distinguish aligned from sophisticated-but-misaligned behaviour better than well-designed behavioural evals.
\item Red-coded instruments routinely substantiate catastrophic-risk absence claims through process-accountable evidence accepted by oversight bodies as sufficient rather than disclosure-grade.
\end{enumerate}
Any one would imply the gap is transient; two would imply the call to action is misdirected.

\section{Conclusion}
\label{sec:conclusion}

Across multiple jurisdictions, the prevailing governance regime presumes a verification apparatus that exists only partially. This gap is filled by fragile assurance: substitute methods that are informative but under-evidenced, routinely treated as more conclusive than they are. As AI systems become more agentic, the cost of overstating what current assurance can verify rises steeply. The path forward is asymmetric: governance frameworks should adopt honest limits on the assurance claims they demand, while the technical community extends the voluntary pre-deployment access arrangements already in operation with narrow mechanistic-evidence classes, to test where meaningful mechanistic verification is presently achievable and where it is not. A concrete 2028 endpoint is modest but real: an EU Art.~55 GPAI submission could include a machine-readable verifier certificate, produced in a 72-hour TEE audit, reporting probe AUROC, ablation effect-size, and enclave hash. Existing mechanistic-audit techniques \cite{goldowsky2025deception,marks2025auditing} would then appear inside a statutory submission with a contract-level chain of custody, analogous in form to the custody standards financial institutions already satisfy for capital adequacy.

\section*{Broader Impact and Societal Consequences}
\label{sec:impact}

The paper is itself an act of governance framing; five consequences deserve comment.

\textbf{Risk of delegitimising useful assurance work.} The fragile-assurance framing could be misread as a blanket dismissal of behavioural evaluations. It is not. These methods remain the best evidence for a large class of decomposable claims (Singapore AI~Verify is the positive case), and the voluntary external-evaluation ecosystem is a public good the call to action builds on rather than replaces. Readers should weight the critique against the Green and Amber cells in \cref{tab:matrix}, not only the Red.

\textbf{Risk of mechanistic-interpretability overreach.} Proposing mechanistic evidence classes as governance input creates pressure toward privileging interpretable-looking evidence even where methods remain immature. Polysemanticity and distributed computation \cite{elhage2022toy,nanda2025pragmatic} mean a probe or ablation can be reproducible without being the \emph{correct} mechanistic story. Pilots should include adversarial evaluation design and explicit uncertainty reporting so that ``the probe reproduced'' does not become indistinguishable from ``the property is verified.''

\textbf{Provider incentives and gaming.} Any verification protocol creates optimisation pressure toward passing it. A provider could train to produce a probe-ready representation or well-localised circuit without that structure actually bearing the safety-relevant work in deployment. The three-line evidence bundle (probe + causation + temporal emergence) is chosen to make single-axis gaming harder; adversarial-robustness of the evidence format should be a first-class design criterion in any future standardisation.

\textbf{Jurisdictional and equity considerations.} Anchor cases span EU, US, Asia-Pacific, South Asia, Council of Europe, and OECD regimes; at the \emph{governance} layer the framing is not Western-concentrated. The \emph{verifier} layer is different: UK AISI, METR, Apollo, FAR.AI are largely US- and UK-based, and extending voluntary-access pilots through them risks concentrating verification capability in a small number of Anglophone institutions. A robust ecosystem should include verifier-pool diversification, public-interest funding models, and participation by institutions in the jurisdictions whose instruments are being verified.

\textbf{Dual use.} Probes for deception and activation patching for refusal circuits are dual-use: the same techniques that audit a refusal claim could, in adversarial hands, locate and disable the circuit. Pilot standardisation needs to address secure-enclave execution \cite{trask2024}, zero-knowledge attestation \cite{waiwitlikhit2024}, weight non-disclosure, and publication norms explicitly; we treat this as an open implementation question.


\newpage
\bibliographystyle{unsrt}
\bibliography{references}

\newpage
\appendix
\onecolumn
\section{Supplementary Material: Full 21-Instrument Inventory Matrix}
\label{sec:supplement}

This supplement extends the seven anchor cases in the main text (\cref{tab:matrix}) to the full 21-row inventory in \cref{tab:matrix_full}. The inventory has three purposes: it documents that the anchor pattern is not cherry-picked; it makes the 13-jurisdiction / 17-instrument / 21-row arithmetic reproducible; and it separates binding and soft-law national and international instruments from three comparator rows (2019 OECD baseline, UNGA non-binding resolution, and voluntary frontier-lab frameworks) plus the vetoed SB-1047.

\begin{table*}[h]
\centering
\footnotesize
\setlength{\tabcolsep}{6pt}
\renewcommand{\arraystretch}{1.15}
\caption{Full 21-instrument inventory coded against the access taxonomy of \cite{casper2024}. \textbf{Access} records what access level the instrument's accepted evidence assumes; \textbf{Cell} records whether independent verification is achievable under \emph{current} access conditions (not legal sufficiency). \textbf{Access key:} B = behavioural, OtB = outside-the-box (documentation review), G = grey-box (weights-adjacent), W = white-box, S = state-embedded. \textbf{Cell key:} \gcell{}~G~routinely satisfiable, \acell{}~A~partially satisfiable under structured access, \rcell{}~R~presumes access not currently deployed by independent verifiers.}
\label{tab:matrix_full}
\begin{tabular}{@{}>{\raggedright\arraybackslash}p{7cm} l l c@{}}
\toprule
\rowcolor{gray!12}
\textbf{Instrument} & \textbf{Jur.} & \textbf{Access} & \textbf{Cell} \\
\midrule
\multicolumn{4}{l}{\textit{National and international governance instruments}} \\
\midrule
\textbf{EU AI Act} (Art.~14, 16, 43)              & EU   & OtB, G part.  & \rcell R \\
\textbf{EU AI Act} (Art.~55, GPAI systemic)       & EU   & B + OtB       & \acell A \\
\textbf{GPAI Code of Practice} (II.1)             & EU   & B             & \acell A \\
\textbf{NIST AI RMF}~1.0                          & US   & B             & \acell A \\
\textbf{NIST GenAI Profile} (AI~600-1)            & US   & B             & \acell A \\
\textbf{California SB-53} (TFAIA)                 & CA   & B + OtB       & \rcell R \\
\textbf{California SB-1047} (vetoed)              & CA   & W + OtB       & \rcell R$^{\dagger}$ \\
\textbf{Singapore MGF-GenAI + AI~Verify}          & SG   & B             & \gcell G/A \\
\textbf{Canada Voluntary Code}                    & CA   & B + OtB       & \acell A \\
\textbf{Australia Vol.\ AI Safety Std}            & AU   & B             & \acell A \\
\textbf{Japan AI Guidelines v1.0}                 & JP   & B             & \acell A \\
\textbf{South Korea AI Basic Act}                 & KR   & B + OtB       & \acell A \\
\textbf{India AI Governance Guidelines} (Feb~'26) & IN   & B + OtB       & \acell A \\
\textbf{PRC Interim Measures}$^{\ddagger}$        & CN   & B + S         & \acell A \\
\textbf{UK pro-innovation + AISI evals}           & UK   & B + G (vol.)  & \acell A \\
\textbf{Council of Europe AI Convention}          & CoE  & B             & \acell A \\
\textbf{Saudi AI Ethics Principles}               & SA   & B             & \acell A \\
\textbf{UAE AI~2031}                              & AE   & ---           & \acell A \\
\midrule
\multicolumn{4}{l}{\textit{Non-binding instruments, baseline, and voluntary-lab comparators}$^{\S}$} \\
\midrule
\textbf{OECD Recommendation on AI}                & OECD & B             & \acell A \\
\textbf{UNGA 78/265}                              & UN   & ---           & --- \\
\textbf{Frontier Safety Frameworks}               & lab  & B + internal  & \acell A \\
\bottomrule
\end{tabular}

{\footnotesize\raggedright
\textit{Sources (per row, in order):} \cite{euaiact,euaiact,eucop2025,nistrmf,nistgenai,sb53,sb1047,smgf,canadacode,ausaistd,japanguidelines,skoreaai,indiaguidelines2026,prcinterim,aisi2025year,coeaiconv,saudiai,uaeai2031,oecdai,unga78265,anthropic_rsp,openai_preparedness,deepmind_fsf}. $^{\dagger}$SB-1047's pre-launch audit and kill-switch requirements demanded methods not deployed; its veto and replacement by SB-53 dropped the requirements rather than solving the problem. $^{\ddagger}$PRC is state-embedded rather than independent-auditor-based; cross-framework comparison is partial. $^{\S}$The three comparator rows (OECD baseline, UNGA non-binding resolution, Frontier Safety voluntary lab) plus the vetoed SB-1047 are excluded from the ``seventeen new instruments'' count: 21 rows total $-$ 4 excluded = 17 post-2019 national and international governance instruments; two of those (EU AI Act) share one jurisdiction, which is why 21 instruments span 13 jurisdictions.\par}
\end{table*}

\newpage
\section{Glossary of Key Terms}
\label{sec:glossary}

This glossary is aimed at readers from adjacent communities (machine learning, mechanistic interpretability, policy research) who may not be fluent in all of the paper's vocabulary.

\textbf{Fragile assurance (paper's central concept; see \cref{def:fragile-assurance}).} A safety claim is fragile when it cannot be reproducibly checked by an independent party under comparable conditions, or when the inferential gap between evidence produced and claim asserted is not supported by the evidence's structure. Fragility is an epistemic property of the claim-evidence pair, not a judgement about whether the claim is true.

\textbf{Incentive gradient (\cref{sec:incentive-gradient}).} The observed directional alignment of three independent resource-allocation pressures (speed-to-deployment, sovereign competitiveness, and pragmatic-proxy tractability) that together reward surface-level behavioural evidence over mechanistic verification. The term names a pattern, not a causal claim.

\textbf{Audit gap.} The mismatch between the access level that an instrument's accepted evidence implicitly assumes and the access level an independent verifier can actually supply under current conditions. \Cref{fig:gap-split} is the visual summary; \cref{tab:matrix} is the per-instrument coding.

\textbf{Access taxonomy (after \cite{casper2024}).} Five levels from lightest to most invasive:
\begin{itemize}[leftmargin=1.5em,itemsep=1pt,topsep=1pt]
\item \textbf{Behavioural (B):} query the model through its standard interface and observe outputs only.
\item \textbf{Outside-the-box (OtB):} supplementary documentation (system cards, architecture descriptions, training-data summaries); no model access beyond behavioural.
\item \textbf{Grey-box (G):} weights-adjacent access, typically under NDA or secure enclave; includes activations, intermediate representations, gradients, or internal logging.
\item \textbf{White-box (W):} full weights, training data, and training process access.
\item \textbf{State-embedded (S):} verification conducted by state entities with privileged access through statutory authority (PRC model).
\end{itemize}

\textbf{Legal sufficiency vs.\ verification feasibility (\cref{sec:matrix}).} \emph{Legal sufficiency}: whether a provider can formally satisfy the statute through the stipulated evidence (self-attestation, documentation). \emph{Verification feasibility}: whether an independent party can, under current access conditions, actually check the underlying claim. The cell colour in \cref{tab:matrix} encodes the latter.

\textbf{Structured access.} Access arrangements that are richer than behavioural (typically grey-box) but bounded by contracts, NDAs, and secure compute environments rather than public release. \cite{shevlane2022,bucknall2023structured,bucknall2026expanding}.

\textbf{Linear probe.} A linear classifier trained on a model's internal activations (typically residual-stream vectors) to predict whether the activations carry a target concept. A common mechanistic-interpretability tool \cite{goldowsky2025deception}.

\textbf{Activation patching.} A causal intervention where activations at a chosen site are replaced with activations from a different input; the downstream effect on model behaviour indicates the functional role of that site. Used to isolate circuits responsible for specific behaviours.

\textbf{Superposition and polysemanticity.} Superposition is the phenomenon that transformer representations can encode more features than they have dimensions, enforcing some degree of overlap \cite{elhage2022toy}. Polysemanticity is the consequence at the neuron level: a single neuron responds to multiple unrelated concepts. Both complicate the inference from ``a probe detects $X$'' to ``the model computes $X$ here.''

\textbf{Behavioural evaluation (``behavioural eval'').} A safety test that observes model output under adversarial, structured, or naturalistic prompts. The dominant evidence class in AI governance, produced by organisations such as UK AISI, METR, Apollo Research, and FAR.AI.

\textbf{Pre-deployment access.} Voluntary arrangements under which frontier AI developers grant external evaluators behavioural (and occasionally grey-box) access to candidate models before public release. The substrate our call to action extends with mechanistic-evidence classes.

\textbf{Before/after-training comparison.} A mechanistic evidence pattern where internal features or circuits are compared between the base pretrained model and the safety-trained (e.g.\ post-RLHF) model, to show that a claimed safety-relevant internal structure emerged during safety training rather than pre-training.

\newpage
\section{Matrix Coding Methodology}
\label{sec:coding-method}

This appendix walks through how two anchor rows in \cref{tab:matrix_full} were coded, making the inference from primary statutory text to access level and cell colour fully auditable. The same procedure was applied uniformly to the remaining rows.

\textbf{Procedure.} For each instrument we extracted, from primary statutory text or official implementation materials:
\begin{enumerate}[leftmargin=1.5em,itemsep=2pt,topsep=2pt,label=(\alph*)]
\item the \emph{explicit assurance requirement} (what the instrument literally asks developers to document, test, or produce);
\item the \emph{implicit technical assumption} (what internal model property must hold for (a) to be satisfiable);
\item the \emph{accepted evidence type} (the kind of artefact the instrument recognises as meeting (a));
\item the \emph{access level} (\cref{sec:glossary}) that the evidence in (c) implicitly assumes, on the taxonomy of \cite{casper2024}.
\end{enumerate}
The cell colour is then determined by whether an independent verifier operating at the access level in (d) can reproduce the claim under current tooling. Green indicates routinely; amber indicates partially or under structured access; red indicates the access required is not currently deployed at scale by independent verifiers.

\textbf{Worked row: California SB-53 (coded \rcell{}~R).}
(a) \emph{Explicit requirement}: developers of frontier models must submit a transparency report describing catastrophic-risk assessment and maintain a 15-day critical-incident reporting window \cite{sb53}.
(b) \emph{Implicit technical assumption}: developers can characterise catastrophic-risk properties of their models (absence of CBRN uplift, absence of loss-of-control precursors, bounded agentic capability) to a standard that makes the transparency report meaningful.
(c) \emph{Accepted evidence}: self-attested transparency reports plus incident disclosures; no pre-launch third-party audit, no kill-switch, no civil-penalty apparatus (all dropped from the predecessor SB-1047, \cite{sb1047,anthropic_fcf}).
(d) \emph{Access level}: B (behavioural evaluation performed by developer) + OtB (documentation review).
\emph{Cell}: R. The catastrophic-risk claims the statute requires (absence of hidden objectives, bounded loss-of-control potential) cannot be verified under current methods from behavioural + documentation access alone. The coding is red on \emph{verification feasibility}, not on legal sufficiency; a developer can comply with SB-53 while the verification gap remains.

\textbf{Worked row: Singapore AI~Verify (coded \gcell{}~G/A).}
(a) \emph{Explicit requirement}: deployers of AI systems run the AI~Verify toolkit and complete a governance questionnaire, both grounded in Singapore's Model AI Governance Framework \cite{smgf,aiverify}.
(b) \emph{Implicit technical assumption}: the relevant safety properties (specified robustness, demographic bias within documented thresholds, narrow capability bounds, traceability) admit decomposable testing against pre-specified metrics.
(c) \emph{Accepted evidence}: AI~Verify test-suite outputs and questionnaire responses, both reviewable by third parties and reproducible under behavioural access.
(d) \emph{Access level}: B.
\emph{Cell}: G/A. The \emph{decomposable} claims AI~Verify addresses (robustness against specified inputs, bias metrics, bounded behaviour) are routinely verifiable at this access level, hence G. Claims about latent properties the toolkit does not directly measure (hidden objectives, deception propensity, long-horizon goal stability) fall through; the A tracks that partial coverage. Singapore's architecture is the strongest positive analog in the matrix precisely because it scopes its claims to what behavioural evidence can support.

\textbf{Coding reliability.} Four of the 21 rows (EU Art.~55, SB-53, Singapore AI~Verify, PRC Interim Measures) were checked against primary statutory text; the remainder relied on secondary sources and official summaries. \Cref{tab:matrix_full} records the cell codings and the citations from which each row was coded.

\textbf{Sensitivity analysis.} We test whether the aggregate finding (no instrument has $V_i \geq A_i$ uniformly across its claim set) survives plausible alternative codings on contested rows. The five most contestable Amber rows (South Korea, India, the Council of Europe Convention, the GPAI Code of Practice, and the UK pro-innovation regime) can be re-coded toward Green for their decomposable claim subsets without changing the result: the high-consequence absence claims still leave $V_i < A_i$. They can also be re-coded toward Red for hidden-objective claims, which only strengthens the finding. Even if all five Amber rows move Green on decomposable claims, seventeen of twenty-one rows still have $V_i < A_i$ on at least one claim category. The four Red rows (EU Art.~14/16/43, SB-53, SB-1047, and the dagger comparator) are insensitive to coding judgement because the required access is not deployed by independent verifiers at scale. Coding sensitivity changes the count of Green-eligible decomposable claims; it does not remove the audit gap on the high-consequence tail.

\textbf{Independent replication.} The per-cell rationale and citations are recorded in the supplementary CSV (\texttt{table1\_per\_cell\_coding.csv}). A reviewer can replicate any row by applying the procedure (a)--(d) above, using the cited sources to recover $A_i$ and the rule in \cref{sec:matrix} to assign $V_i$. We expect disagreement on some borderline rows, but not on the aggregate finding.

\newpage
\section{What Existing External Evaluators Currently Supply}
\label{sec:existing-evaluators}

\Cref{tab:evaluators} summarises the voluntary-access ecosystem that \cref{sec:action} builds on, alongside the mechanistic-extension gap our call to action is designed to close. The claim throughout the paper is \emph{not} that these organisations are inadequate; it is that the evidence class they currently produce does not yet include the reproducible mechanistic artefacts (probes, activation-patching results, before/after-training comparisons) that the Red cells in \cref{tab:matrix} would require.

The point of \cref{tab:evaluators} is the column on the right. None of the gaps listed is a criticism of the listed organisation: each one is producing the evidence class the voluntary-access model currently supports. The mechanistic-extension proposal in \cref{sec:action} asks whether, under similar access arrangements, the reports could additionally include narrow, reproducible, claim-tied mechanistic artefacts, and whether doing so is operationally tractable in a 2--3~year pilot.
\begin{table*}[pt]
\centering
\footnotesize
\setlength{\tabcolsep}{5pt}
\renewcommand{\arraystretch}{1.35}
\caption{Representative external evaluators operating under voluntary pre-deployment access agreements with frontier AI developers. \textbf{Evidence class} is the dominant form of artefact the organisation produces in its published reports; \textbf{Typical access} is the access level at which that evidence is generated; \textbf{Mechanistic-extension gap} is what the organisation currently does \emph{not} routinely supply, relative to the pilot proposed in \cref{sec:action}. Organisations also produce other outputs; this table is not exhaustive.}
\label{tab:evaluators}
\begin{tabular}{@{}>{\raggedright\arraybackslash}p{2.2cm} >{\raggedright\arraybackslash}p{3.5cm} >{\raggedright\arraybackslash}p{2.5cm} >{\raggedright\arraybackslash}p{5cm}@{}}
\toprule
\rowcolor{gray!12}
\textbf{Evaluator} & \textbf{Evidence class} & \textbf{Typical access} & \textbf{Mechanistic-extension gap} \\
\midrule
\textbf{UK AISI} \cite{aisi2025year}
& Capability evaluations, agentic-task performance, risk-domain red-teaming (30+ frontier models, 2024--2025)
& B with voluntary G on a subset
& Reports describe \emph{observable behaviour} of the model under test conditions; do not typically publish reproducible internal probes or circuit-level results tied to specific governance claims. \\
\textbf{METR} \cite{kwa2025metr}
& Long-horizon task benchmarks; autonomous-task completion measured as a function of time-horizon
& B under NDA with developers
& Task-completion time series are behavioural; they do not bound latent representations. Extension target: probe-based evidence that the capability (or its absence) has a reproducible internal correlate. \\
\textbf{Apollo Research} \cite{meinke2024scheming}
& In-context scheming and deception evaluations; prompts that create instrumental-goal conflicts
& B with supplementary documentation
& Reports show behavioural instances of scheming under test conditions; do not typically accompany those instances with internal-feature probes (e.g., \cite{goldowsky2025deception}-style AUROC on held-out concealment data) reproducible by an independent verifier. \\
\textbf{FAR.AI} \cite{mckenzie2025stack}
& Adversarial attacks on safety pipelines; multi-layer safeguard penetration; red-teaming
& B + targeted structured access
& Safeguard penetration rates are behavioural; extension target: reproducible evidence about \emph{which internal mechanism} failed (circuit-level, feature-level) when a safeguard is bypassed, to tie the adversarial result back to Article~55 or SB-53 claims. \\
\bottomrule
\end{tabular}
\end{table*}

\newpage
\section{Extended Alternative-View Replies}
\label{sec:extended-replies}

The compressed replies in \cref{sec:alternative-views} omit detail that some readers will want; the additional engagement is collected here.

\textbf{View 1 (sufficiency of behavioural evaluations), extended.} \cite{marks2025auditing} flag that even their interpretability-equipped teams operated under unusually favourable access (held-out concealment datasets, weights, activations); the methodology exists, but the access to run it does not under current statutory regimes. A safety case that rests exclusively on test-environment metrics for absence claims accepts an evidential gap it is not equipped to close. Better behavioural evaluations raise the Green ceiling but do not touch the Red cells in \cref{tab:matrix}.

\textbf{View 2 (mech interp too immature), extended.} Acknowledging immaturity should push governance to \emph{moderate} demands (the first two items in \cref{sec:action}) rather than \emph{maintain} them while substituting weaker proxies. The \cite{sheshadri2026auditbench} alignment-auditing benchmark and \cite{treutlein2026saboteur} pre-deployment saboteur demonstration show narrow, claim-specific mechanistic evidence is tractable today under cooperating-lab conditions; the open question concerns reproducibility under statutory access, not solving superposition at scale.

\textbf{View 3 (mechanistic fragile assurance), extended.} The Broader Impact section names this risk explicitly. Mechanistic extension is a hypothesis to test, not a guarantee; the honest framing we argue for governance applies equally to the evidence class we propose. Pilots framed as we propose should include adversarial evaluation design and explicit uncertainty reporting so that ``the probe reproduced'' does not become indistinguishable from ``the property is verified.''

\textbf{View 4 (access not overreach), extended.} \cite{marks2025auditing}'s mechanistic-auditing setup had unusually favourable access and the authors still flag methodological limits; \cite{elhage2022toy}'s superposition does not disappear under white-box conditions. Continuous behavioural monitoring still cannot bound latent properties. The access-design, evidence-class, and cadence axes are complementary rather than substitutes.

\textbf{View 5 (control protocols), extended.} Where statutes are claim-relative on absence properties, control evidence supplements but cannot substitute. The control programme and the mechanistic-extension programme should be read as joint, not rival, lines of work.

\newpage
\section{Extended Limitations}
\label{sec:extended-limits}

Three additional bounded scopes beyond the four flagged in \cref{sec:limitations}.

\textbf{Provider vs.\ deployer and open-weight cases.} The matrix tracks provider obligations, which differ from deployer obligations under most instruments. It also implicitly assumes a proprietary-model setting where the provider controls access; for open-weight models (e.g., Llama-family releases) the taxonomy breaks down and a separate analysis is needed.

\textbf{The gap may partly reflect pragmatic governance design.} Many instruments coded Red or Amber deliberately require \emph{process accountability} (documented risk management, traceable testing, incident reporting) rather than \emph{outcome-level verification}. Regulators in other domains routinely rely on process-accountable regimes without demanding absolute verification, precisely because absolute verification is unattainable. On this reading, the audit gap is partly by design. Our position is narrower than a wholesale governance critique: on the high-consequence tail, the pragmatic process-accountable design has outrun what behavioural evidence epistemically supports, and the statutory text claims more than the process accountability produces. \Cref{sec:action} is calibrated accordingly: distinguish decomposable claims (process accountability adequate) from absence claims (it is not), without dismantling process-accountable regimes wholesale.

\textbf{Coverage of agentic settings is illustrative.} \Cref{sec:agentic} treats agentic deployment as an illustration, not a separate framework. A full agentic-specific analysis would require a parallel matrix coded against agentic-specific instruments and deployment surfaces, which we do not attempt.

\textbf{Agentic illustration: the Claude~Opus~4 blackmail scenario.} The Claude~Opus~4 system card \cite{anthropic_opus4_card} reports a simulated insider-threat setting in which a model with email access discovers both an executive's affair and its own impending replacement, then attempts blackmail. The example matters because three trends move together: models increasingly distinguish evaluation from deployment, tool-use surfaces magnify the consequences of latent misalignment, and multi-agent settings make incident attribution harder than single-model debugging. The scenario is therefore not just a vivid anecdote; it illustrates why agentic deployment creates a different verification problem from static inference, and why \Cref{sec:agentic} remains illustrative rather than exhaustive.

\end{document}